\documentclass[runningheads]{llncs}
\usepackage{graphicx}
\usepackage{amsmath,amssymb} 
\usepackage{color}
\usepackage{multirow}
\usepackage[width=122mm,left=12mm,paperwidth=146mm,height=193mm,top=12mm,paperheight=217mm]{geometry}

\def\ie{\emph{i.e.}}
\def\eg{\emph{e.g.}}

\def\etal{\emph{et al.}}

\newif\ifsqueeze
\squeezetrue  
\ifsqueeze
  \newcommand{\Caption}[1]{\vspace{-2.5mm}\caption{{\footnotesize #1}}}
  \newcommand{\Section}[1]{\vspace{-1mm} \section{#1} \vspace{-2mm}}
  
  \newcommand{\Subsection}[1]{\vspace{-1mm} \subsection{#1} \vspace{-1mm} }

\else
  \newcommand{\Caption}[1]{\caption{#1}}
  \newcommand{\Section}[1]{\section{#1}}
  
  \newcommand{\Subsection}[1]{\subsection{#1}}

\fi

\begin{document}

\pagestyle{headings}
\mainmatter

\title{Recognition in Terra Incognita}

\titlerunning{Recognition in Terra Incognita}

\authorrunning{S. Beery, G. Van Horn, and P. Perona}

\author{Sara Beery, Grant Van Horn, and Pietro Perona}
\institute{Caltech\\
	\email{ \{sbeery,gvanhorn,perona\}@caltech.edu}
}

\maketitle
\begin{abstract}
It is desirable for detection and classification algorithms to generalize to unfamiliar environments, but suitable benchmarks for quantitatively studying this phenomenon are not yet available. We present a dataset designed to measure recognition generalization to novel environments. The images in our dataset are harvested from twenty camera traps deployed to monitor animal populations. Camera traps are fixed at one location, hence the background changes little across images; capture is triggered automatically, hence there is no human bias. The challenge is learning recognition in a handful of locations, and generalizing animal detection and classification to new locations where no training data is available. In our experiments state-of-the-art algorithms show excellent performance when tested at the same location where they were trained. However, we find that generalization to new locations is poor, especially for classification systems.\footnote{The dataset is available at   \url{https://beerys.github.io/CaltechCameraTraps/}}
\keywords{Recognition, transfer learning, domain adaptation, context, dataset, benchmark.}
\end{abstract}
\section{Introduction}
Automated visual recognition algorithms have recently achieved human expert performance at visual classification tasks in field biology~\cite{van2017inaturalist,norouzzadeh2017automatically,merlinBirdID} and medicine~\cite{esteva2017dermatologist,poplin2018prediction}. Thanks to the combination of deep learning~\cite{fukushima1982neocognitron,lecun1998gradient}, Moore's law~\cite{schaller1997moore} and very large annotated datasets~\cite{imagenet_cvpr09,lin2014microsoft} enormous progress has been made during the past 10 years. Indeed, 2017 may come to be remembered as the year when automated visual categorization surpassed human performance.

\begin{figure}[t]
\begin{minipage}[b]{.3\linewidth}
  \centering
  \centerline{\includegraphics[width=4.0cm]{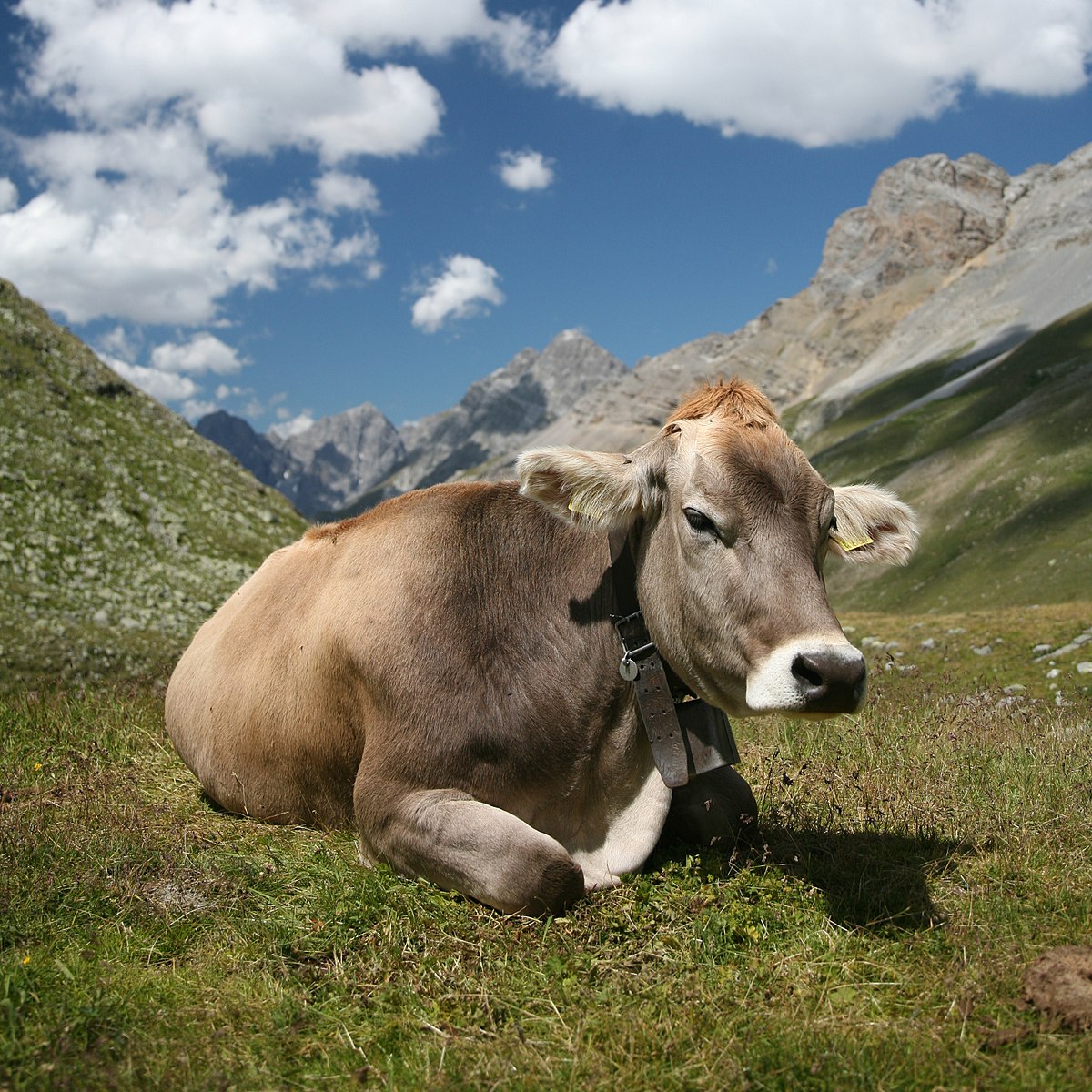}}
{\scriptsize (A) {\bf Cow: 0.99},
Pasture: 0.99,
Grass: 0.99,
No Person: 0.98,
Mammal: 0.98}

  \medskip
\end{minipage}
\hfill
\begin{minipage}[b]{0.3\linewidth}
  \centering
  \centerline{\includegraphics[width=4.0cm]{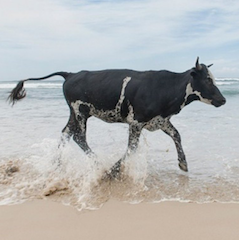}}
{\scriptsize (B) No Person: 0.99,
Water: 0.98,
Beach: 0.97,
Outdoors: 0.97,
Seashore: 0.97}

  \medskip
\end{minipage}
\hfill
\begin{minipage}[b]{0.3\linewidth}
  \centering
  \centerline{\includegraphics[width=4.0cm]{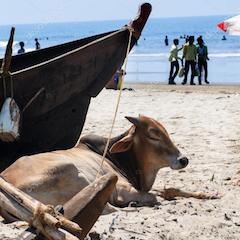}}
{\scriptsize (C) No Person: 0.97,
{\bf Mammal: 0.96},
Water: 0.94,
Beach: 0.94,
Two: 0.94}
  \medskip
\end{minipage}
\vspace{-0.15in}
\setlength{\belowcaptionskip}{-10pt}
\caption{{\bf Recognition algorithms generalize poorly to new environments.} Cows in `common' contexts (e.g. Alpine pastures) are detected and classified correctly (A), while cows in uncommon contexts (beach, waves and boat) are not detected (B) or classified poorly (C). Top five labels and confidence produced by ClarifAI.com shown.}
\label{fig-generalization}
\end{figure}

However, it is known that current learning algorithms are dramatically less data-efficient than humans~\cite{van2017devil}, transfer learning is difficult~\cite{pan2010survey}, and, anecdotally, vision algorithms do not generalize well across datasets~\cite{torralba2011unbiased,welinder2013lazy} (Fig.~\ref{fig-generalization}). These observations suggest that current algorithms rely mostly on rote pattern-matching, rather than abstracting from the training set `visual concepts'~\cite{murphy2004big} that can generalize well to novel situations. In order to make progress we need datasets that support a careful analysis of generalization, dissecting the challenges in detection and classification: variation in lighting, viewpoint, shape, photographer's choice and style, context/background. Here we focus on the latter: generalization to new environments, which includes background and overall lighting conditions.

Applications where the ability to generalize visual recognition to new environments is crucial include surveillance, security, environmental monitoring, assisted living, home automation, automated exploration (\eg\ sending rovers to other planets). Environmental monitoring by means of camera traps is a paradigmatic application. Camera traps are heat- or motion-activated cameras placed in the wild to monitor and investigate animal populations and behavior. Camera traps have become inexpensive, hence hundreds of them are often deployed for a given study, generating a deluge of images. Automated detection and classification of animals in images is a necessity. The challenge is training animal detectors and classifiers from data coming from a few pilot locations such that these detectors and classifiers will generalize to new locations. Camera trap data is controlled for environment including lighting (the cameras are static, and lighting changes systematically according to time and weather conditions), and eliminates photographer bias (the cameras are activated automatically). 

Camera traps are not new to the computer vision community \\\cite{ren2013ensemble,yu2013automated,wilber2013animal,chen2014deep,lin2014foreground,swanson2015snapshot,zhang2015coupled,zhang2016animal,miguel2016finding,giraldo2017camera,yousif2017fast,villa2017towards,norouzzadeh2017automatically}. Our work is the first to identify camera traps as a unique opportunity to study generalization, and we offer the first study of generalization to new environments in this controlled setting. We make here three contributions: (a) a novel, well-annotated dataset to study visual generalization across locations, (b) a benchmark to measure algorithms' performance, and (c) baseline experiments establishing the state of the art. Our aim is to complement current datasets utilized by the vision community for detection and classification\cite{imagenet_cvpr09,lin2014microsoft,everingham2010pascal,openimages} by introducing a new dataset and experimental protocol that can be used to systematically evaluate the generalization behavior of algorithms to novel environments. In this work we benchmark the current state-of-the-art detection and classification pipelines and find that there is much room for improvement.

\section{Related Work}
\subsection{Datasets}
The ImageNet \cite{imagenet_cvpr09}, MS-COCO \cite{lin2014microsoft}, PascalVOC \cite{everingham2010pascal}, and Open Images \cite{openimages} datasets are commonly used for benchmarking classification and detection algorithms. Images in these datasets were collected in different locations by different people, which enables algorithms to average over photographer style and irrelevant background clutter. However, as demonstrated in Fig.~\ref{fig-generalization}, the context can be strongly biased. Human photographers are biased towards well-lit, well-focused images where the subjects are centered in the frame~\cite{ponce2006dataset,spain2008some}. Furthermore, the number of images per class is balanced, unlike what happens in the real world~\cite{van2017devil}. 

Natural world datasets such as the iNaturalist dataset \cite{van2017inaturalist}, CUB200 \cite{wah2011caltech}, Oxford Flowers \cite{Nilsback06}, LeafSnap \cite{leafsnap_eccv2012}, and NABirds700 \cite{van2015building} are focused on fine-grained species classification and detection. Most images in these datasets are taken by humans under relatively good lighting conditions, though iNaturalist does contain human-selected camera trap images. Many of these datasets exhibit real-world long-tailed distributions, but in all cases there is a large amount of diversity in location and perspective. 

The Snapshot Serengeti dataset~\cite{swanson2015snapshot} is a large, multi-year camera trap dataset collected at $225$ locations in a small region of the African savanna. It is the single largest-scale camera trap dataset ever collected, with over $3$ million images. However, it is not yet suitable for controlled experiments. This dataset was collected from camera traps that fire in sequences of $3$ for each motion trigger, and provides species annotation for groups of images based on a time threshold. This means that sometimes a single species annotation is provided for up to 10 frames, when in fact the animal was present in only a few of those frames (no bounding boxes are provided). Not all camera trap projects are structured in a similar way, and many cameras take shorter sequences or even single images on each trigger. In order to find a solution that works for new locations regardless of the camera trap parameters, it is important to have information about which images in the batch do or do not contain animals. In our dataset we provide annotations on a per-instance basis, with bounding boxes and associated classes for each animal in the frame. 

\subsection{Detection}
Since camera traps are static, detecting animals in the images could be considered either a change detection or foreground detection problem. Detecting changes and/or foreground vs. background in video is a well studied problem \cite{st2015subsense}, \cite{babaee2017deep}. Many of these methods rely on constructing a good background model that updates regularly, and thus degrade rapidly at low frame rates. \cite{zhan2017change} and \cite{benedek2008mixed} consider low frame rate change detection in aerial images, but in these cases there are often very few examples per location. 

Some camera traps collect a short video when triggered instead of a sequence of frames. \cite{lin2014foreground,zhang2016animal,zhang2015coupled} show foreground detection results on camera trap video. Data that comes from most camera traps take sequences of frames at each trigger at a frame rate of $\sim1$ frame per second. This data can be considered ``video," albeit with extremely low, variable frame rate. Statistical methods for background subtraction and foreground segmentation in camera trap image sequences have been previously considered. \cite{ren2013ensemble} demonstrates a graph-cut method that uses background modeling and foreground object saliency to segment foreground in camera trap sequences. \cite{miguel2016finding} creates background models and perform a superpixel-based comparison to determine areas of motion. \cite{giraldo2017camera} uses a multi-layer RPCA-based method applied to day and night sequences. \cite{yousif2017fast} uses several statistical background-modeling approaches as additional signal to improve and speed up deep detection.  These methods rely on a sequence of frames at each trigger to create appropriate background models, which are not always available. None of these methods demonstrate results on locations outside of their training set.

\subsection{Classification}
A few studies tackle classification of camera trap images. \cite{wilber2013animal} showed results classifying squirrels vs. tortoises in the Mojave Desert. \cite{yu2013automated} showed classification results on data that provides image sequences of \~10 frames.  They do not consider the detection problem and instead manually crop the animal from the frame and balance the dataset, resulting in a total of 7,196 images across 18 species with at least 100 examples each. \cite{chen2014deep} were the first to take a deep network approach to camera trap classification, working with data from eMammal \cite{eMammal}.  They first performed detection using the background subtraction method described in \cite{ren2013ensemble}, then classified cropped detected regions, getting 38.31\% top-1 accuracy on 20 common species. \cite{villa2017towards} show classification results on both Snapshot Serengeti and data from jungles in Panama, and saw a boost in classification performance from providing animal segmentations. \cite{norouzzadeh2017automatically} show 94.9\% top-1 accuracy using an ensemble of models for classification on the Snapshot Serengeti dataset. None of the previous works show results on unseen test locations.  
\subsection{Generalization and Domain Adaptation}
Generalizing to a new location is an instance of domain adaptation, where each location represents a domain with its own statistical properties such as types of flora and fauna, species frequency, man-made or other clutter, weather, camera type, and camera orientation. There have been many methods proposed for domain adaptation in classification \cite{csurka2017domain}. \cite{ganin2015unsupervised} proposed a method for unsupervised domain adaptation by maximizing domain classification loss while minimizing loss for classifying the target classes. We generalized this method to multi-domain for our dataset, but did not see any improvement over the baseline. \cite{gebru2017fine} demonstrated results of a similar method for fine-grained classification, using a multi-task setting where the adaptation was from clean web images to real-world images, and \cite{busto2017open} investigated open-set domain adaptation.

Few methods have been proposed for domain adaptation outside of classification. \cite{hoffman2016fcns,chen2017road,zhang2017curriculum} investigate methods of domain adaptation for semantic segmentation, focusing mainly on cars and pedestrians and either adapting from synthetic to real data, from urban to suburban scenes, or from PASCAL to a camera on-board a car. \cite{peng2015learning,tang2012shifting,sun2014virtual,hattori2015learning,xu2014domain} look at methods for adapting detectors from one data source to another, such as from synthetic to real data or from images to video.  Raj, et. al., \cite{raj2015subspace} demonstrated a subspace-based detection method for domain adaptation from PASCAL to COCO.  

\section{The Caltech Camera Traps Dataset}
\begin{figure}
\begin{minipage}[b]{.24\linewidth}
  \centering
  \centerline{\includegraphics[width=3.0cm]{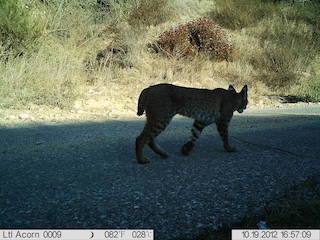}}
  \vspace{.05cm}
\end{minipage}
\hfill
\begin{minipage}[b]{0.24\linewidth}
  \centering
  \centerline{\includegraphics[width=3.0cm]{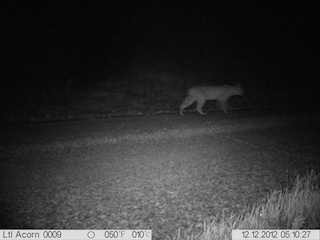}}
\vspace{.05cm}

\end{minipage}
\hfill
\begin{minipage}[b]{.24\linewidth}
  \centering
  \centerline{\includegraphics[width=3.0cm]{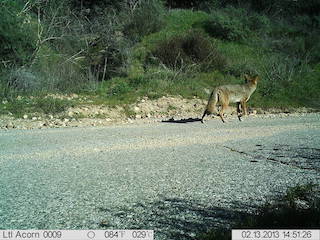}}
\vspace{.05cm}

\end{minipage}
\hfill
\begin{minipage}[b]{.24\linewidth}
  \centering
  \centerline{\includegraphics[width=3.0cm]{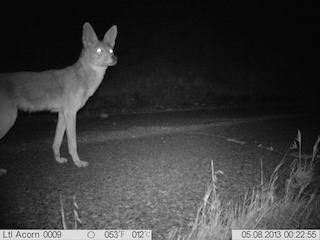}}
\vspace{.05cm}

\end{minipage}

\begin{minipage}[b]{0.24\linewidth}
  \centering
  \centerline{\includegraphics[width=3.0cm]{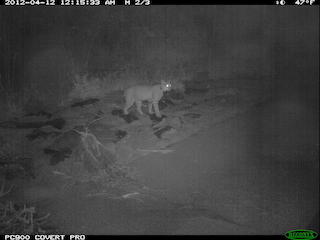}}
\vspace{.05cm}

\end{minipage}
\hfill
\begin{minipage}[b]{.24\linewidth}
  \centering
  \centerline{\includegraphics[width=3.0cm]{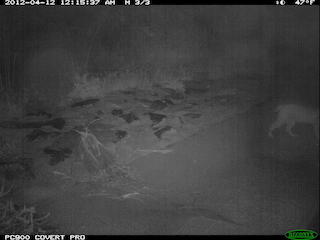}}
\vspace{.05cm}

\end{minipage}
\hfill
\begin{minipage}[b]{0.24\linewidth}
  \centering
  \centerline{\includegraphics[width=3.0cm]{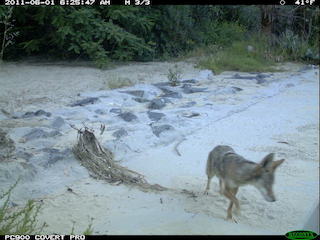}}
\vspace{.05cm}

\end{minipage}
\hfill
\begin{minipage}[b]{.24\linewidth}
  \centering
  \centerline{\includegraphics[width=3.0cm]{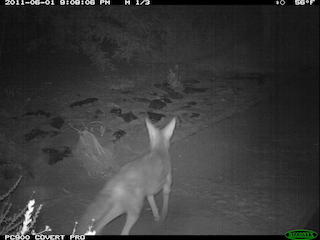}}
\vspace{.05cm}

\end{minipage}

\begin{minipage}[b]{0.24\linewidth}
  \centering
  \centerline{\includegraphics[width=3.0cm]{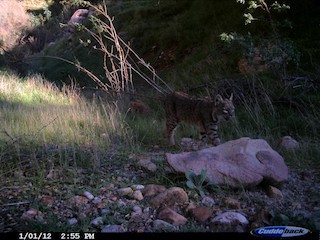}}
\vspace{.05cm}

\end{minipage}
\hfill
\begin{minipage}[b]{.24\linewidth}
  \centering
  \centerline{\includegraphics[width=3.0cm]{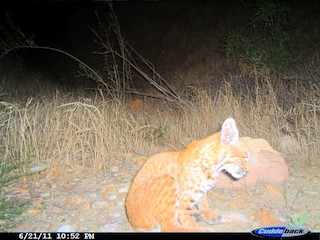}}
\vspace{.05cm}

\end{minipage}
\hfill
\begin{minipage}[b]{0.24\linewidth}
  \centering
  \centerline{\includegraphics[width=3.0cm]{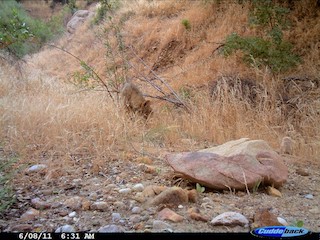}}
\vspace{.05cm}

\end{minipage}
\hfill
\begin{minipage}[b]{.24\linewidth}
  \centering
  \centerline{\includegraphics[width=3.0cm]{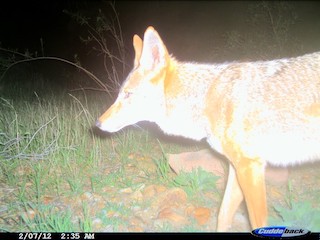}}
\vspace{.05cm}
\end{minipage}
\caption{{\bf Camera trap images from three different locations.} Each row is a different location and a different camera type. The first two cameras use IR, while the third row used white flash.  The first two columns are bobcats, the next two columns are coyotes.}
\label{fig:camtrap_ims}

\begin{minipage}[b]{.3\linewidth}
  \centering
  \centerline{\includegraphics[width=4.0cm]{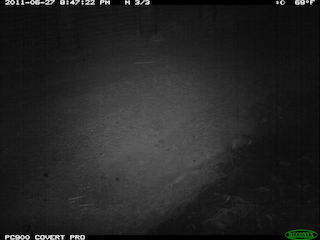}}
  \centerline{(1) Illumination}\medskip
\end{minipage}
\hfill
\begin{minipage}[b]{0.3\linewidth}
  \centering
  \centerline{\includegraphics[width=4.0cm]{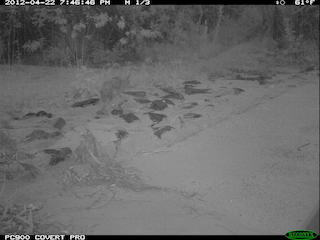}}
  \centerline{(2) Blur}\medskip
\end{minipage}
\hfill
\begin{minipage}[b]{.3\linewidth}
  \centering
  \centerline{\includegraphics[width=4.0cm]{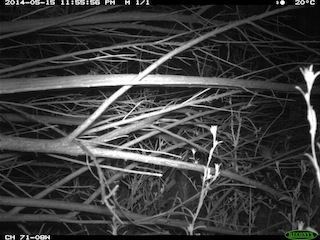}}
  \centerline{(3) ROI Size}\medskip
\end{minipage}

\begin{minipage}[b]{0.3\linewidth}
  \centering
  \centerline{\includegraphics[width=4.0cm]{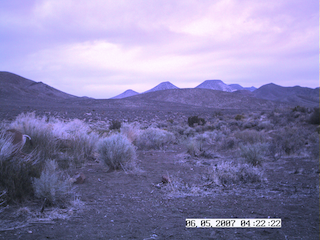}}
  \centerline{(4) Occlusion}\medskip
\end{minipage}
\hfill
\begin{minipage}[b]{.3\linewidth}
  \centering
  \centerline{\includegraphics[width=4.0cm]{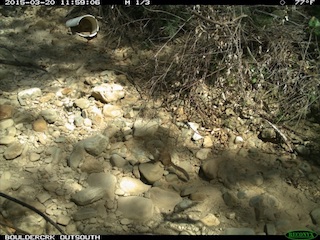}}
  \centerline{(5) Camouflage}\medskip
\end{minipage}
\hfill
\begin{minipage}[b]{0.3\linewidth}
  \centering
  \centerline{\includegraphics[width=4.0cm]{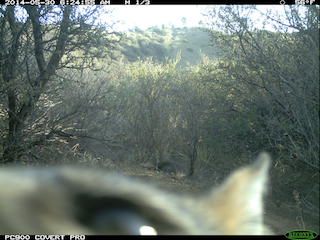}}
  \centerline{(6) Perspective}\medskip
\end{minipage}
\caption{\textbf{Common data challenges}: (1) {\bf Illumination}: Animals are not always salient. (2) {\bf Motion blur}: common with poor illumination at night. (3) {\bf Size of the region of interest} (ROI): Animals can be small or far from the camera. (4) {\bf Occlusion}: e.g. by bushes or rocks. (5) {\bf Camouflage}: decreases saliency in animals' natural habitat. (6) {\bf Perspective}: Animals can be close to the camera, resulting in partial views of the body.}
\label{fig:challenging_ims}
\end{figure}

The Caltech Camera Traps (CCT) dataset contains 243,187 images from 140 camera locations, curated from data provided by the USGS and NPS. Our goal in this paper is to specifically target the problem of generalization in detection and classification. To this end, we have randomly selected 20 camera locations from the American Southwest to study in detail. By limiting the geographic region, the flora and fauna seen across the locations remain consistent. The current task is not to deal with entirely new regions or species, but instead to be able to recognize the same species of animals in the same region with a different camera background. In the future we plan to extend this work to recognizing the same species in new regions, and to the open-set problem of recognizing never-before-seen species. Examples of data from different locations can be seen in Fig.~\ref{fig:camtrap_ims}. 

Camera traps are motion- or heat-triggered cameras that are placed in locations of interest by biologists in order to monitor and study animal populations and behavior. When a camera is triggered, a sequence of images is taken at approximately one frame per second. Our dataset contains sequences of length $1-5$. The cameras are prone to false triggers caused by wind or heat rising from the ground, leading to empty frames. Empty frames can also occur if an animal moves out of the field of view of the camera while the sequence is firing. Once a month, biologists return to the cameras to replace the batteries and change out the memory card. After it has been collected, experts manually sort camera trap data to categorize species and remove empty frames. The time required to sort and label images by hand severely limits data scale and research productivity. We have acquired and further curated a portion of this data to analyze generalization behaviors of state-of-the-art classifiers and detectors. 

\begin{figure}
 \begin{minipage}[b]{.48\linewidth}
  \centering
  \centerline{\includegraphics[height=4cm]{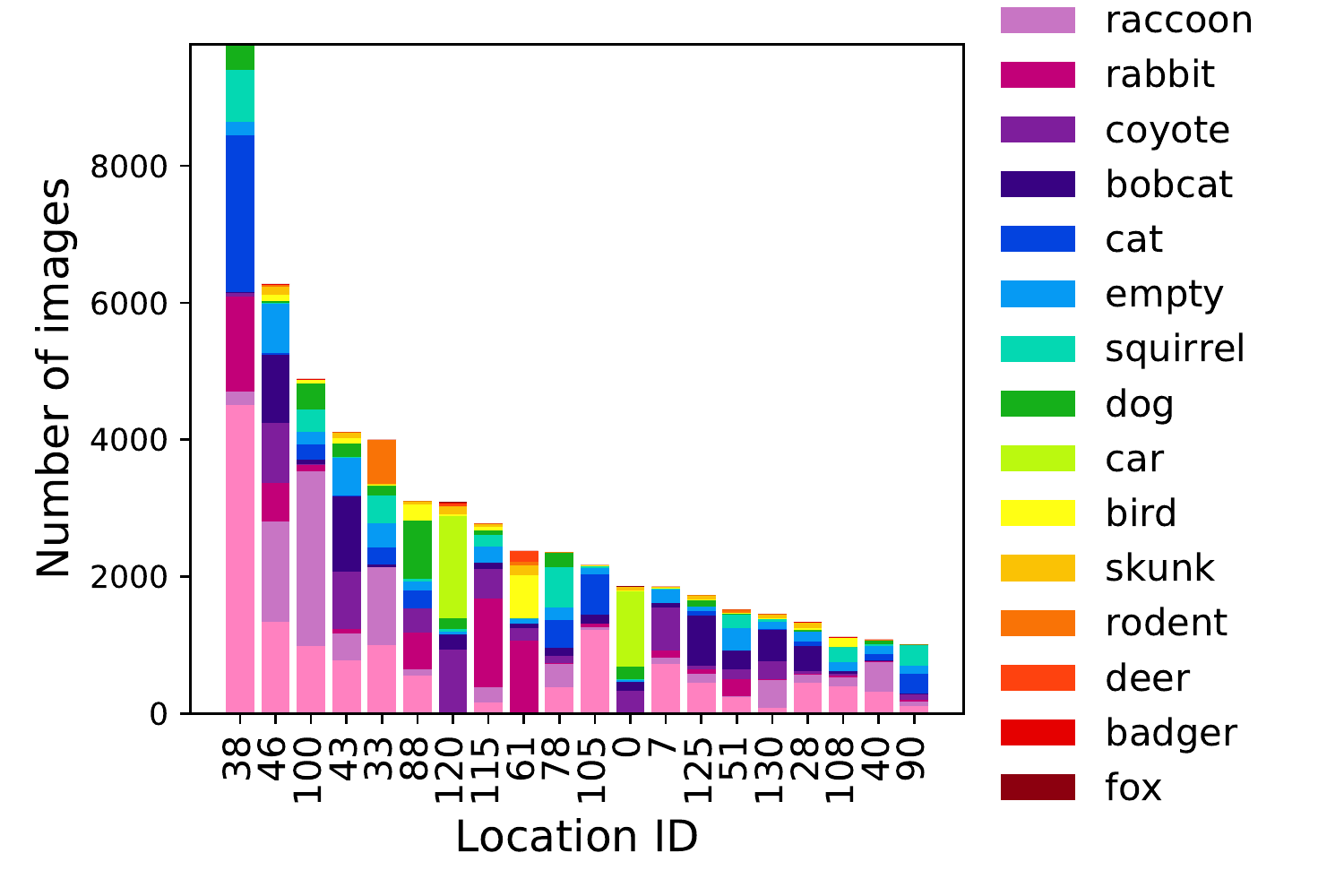}}  
\end{minipage}
\hfill
\begin{minipage}[b]{0.48\linewidth}
  \centering
  \centerline{\includegraphics[height=4.2cm]{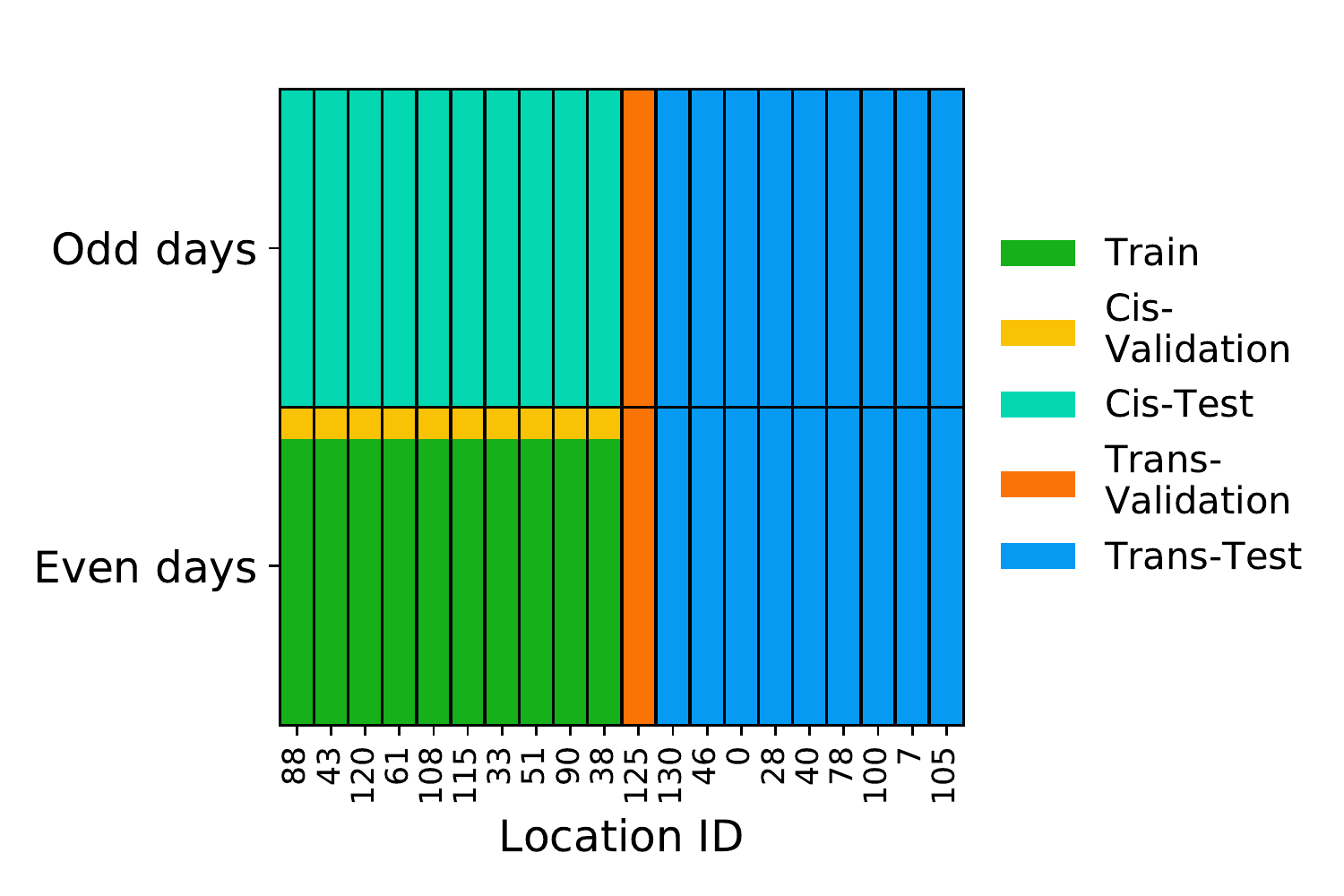}}
\end{minipage}
\caption{(Left) Number of annotations for each location, over 16 classes. The ordering of the classes in the legend is from most to least examples overall. The distribution of the species is long-tailed at each location, and each location has a different and peculiar distribution. (Right) Visualization of data splits. ``Cis" refers to images from locations seen during training, and ``trans" refers to new locations not seen during training.}
\label{fig:annotPerLoc}
\label{fig:datasplit}
\end{figure}

The dataset in this paper, which we call Caltech Camera Traps-20 (CCT-20), consists of $57,868$ images across $20$ locations, each labeled with one of $15$ classes (or marked as empty). Classes are either single species (e.g. "Coyote" or groups of species, e.g. "Bird"). See Fig.~\ref{fig:annotPerLoc} for the distribution of classes and images across locations. We do not filter the stream of images collected by the traps, rather this is the same data that a human biologist currently sifts through. Therefore the data is unbalanced in the number of images per location, distribution of species per location, and distribution of species overall (see Fig.~\ref{fig:annotPerLoc}).  

\subsection{Detection and Labeling Challenges}

The animals in the images can be challenging to detect and classify, even for humans. We have determined that there are six main nuisance factors inherent to camera trap data, which can compound upon each other. Detailed analysis of these challenges can be seen in Fig.~\ref{fig:challenging_ims}. When an image is too difficult to classify on its own, biologists will often refer to an easier image in the same sequence and then track motion by flipping between sequence frames in order to generate a label for each frame (\eg\ is the animal still present or has it gone off the image plane?). We account for this in our experiments by reporting performance at the frame level and at the sequence level. Considering frame level performance allows us to investigate the limits of current models in exceptionally difficult cases.

\subsection{Annotations}
We collected bounding box annotations on Amazon Mechanical Turk, procuring annotations from at least three and up to ten mturkers for each image for redundancy and accuracy.  Workers were asked to draw boxes around all instances of a specific type of animal for each image, determined by what label was given to the sequence by the biologists. We used the crowdsourcing method by Branson \etal~\cite{bboxclustering} to determine ground truth boxes from our collective annotations, and to iteratively collect additional annotations as necessary. We found that bounding box precisions varied based on annotator, and determined that for this data the PascalVOC metric of IoU$ \geq 0.5$ is appropriate for the detection experiments (as opposed to the COCO IoU averaging metric). 

\subsection{Data Split: Cis- and Trans-}
Our goal is exploring generalization to new (i.e. untrained) locations.  Thus, we compare the performance of detection and classification algorithms when they are tested at the same locations where they were trained, vs new locations. For brevity, we refer to locations seen during training as \textit{\bf cis-locations} and locations not seen during training as \textit{\bf trans-locations}.  

From our pool of 20 locations, we selected 9 locations at random to use as trans-location test data, and a single random location to use as trans-location validation data.  From the remaining 10 locations, we use images taken on odd days as cis-location test data.  From within the data taken on even days, we randomly select 5\% to be used as cis-location validation data.  The remaining data is used for training, with the constraint that training and validation sets do not share the same image sequences. This gives us $13,553$ training images, $3,484$ validation and $15,827$ test images from cis-locations, and $1,725$ val and $23,275$ test images from trans-locations. The data split can be visualized in Fig.~\ref{fig:datasplit}. We chose to interleave the cis training and test data by day because we found that using a single date to split the data results in additional generalization challenges due to changing vegetation and animal species distributions across seasons. By interleaving, we reduce noise and provide a clean experimental comparison of results on cis- and trans-locations.

\section{Experiments}
Current state-of-the-art computer vision models for classification and detection are designed to work well on test data whose distribution matches the training distribution. However, in our experiments we are explicitly evaluating the models on a different test distribution. In this situation, it is common practice to employ early stopping~\cite{bengio2012practical} as a means of preventing overfitting to the train distribution. Therefore, for all classification and detection experiments we monitor performance on both the cis- and trans-location validation sets. In each experiment we save two models, one that we expect has the best performance on the trans-location test set (\ie\ a model that generalizes), and one that we expect has the best performance on the cis-location test set (\ie\ a model that performs well on the train distribution).  

\subsection{Classification}
We explore the generalization of classifiers in 2 different settings: full images and cropped bounding boxes. For each setting we also explore the effects of using and ignoring sequence information. Sequence information is utilized in two different ways: \textbf{(1) Most Confident} we consider the sequence to be classified correctly if the most confident prediction from \textit{all} frames grouped together is correct, or \textbf{(2) Oracle} we consider the sequence to be correctly classified if \textit{any} frame is correctly classified. Note that (2) is a more optimistic usage of sequence information. For all classification experiments we use an Inception-v3~\cite{szegedy2016rethinking} model pretrained on ImageNet, with an initial learning rate of 0.0045, rmsprop with a momentum of 0.9, and a square input resolution of 299. We employ random cropping (containing at least 65\% of the region), 
horizontal flipping, and color distortion as data augmentation. 

\setlength{\tabcolsep}{4pt}
\begin{table}
\begin{center}
\caption{Classification top-1 error across experiments. Empty images are removed for these experiments.}
\label{table:results}
\begin{tabular}{l|cc|cc|cc|}
\cline{2-7}
&  \multicolumn{2}{|c|}{Cis-Locations} & \multicolumn{2}{|c|}{Trans-Locations} & \multicolumn{2}{|c|}{Error Increase}\\
\hline
\multicolumn{1}{|c|}{Sequence Information}& Images & Bboxes & Images & Bboxes & Images & Bboxes\\
\hline
\multicolumn{1}{|c|}{None} & 19.06 & 8.14 & 41.04 & 19.56 & 115\% & 140\%\\
\multicolumn{1}{|c|}{Most Confident} & 17.7 & 7.06& 34.53& 15.77 & 95\% & 123\%\\
\multicolumn{1}{|c|}{Oracle}  & 14.92 & 5.52 & 28.69 & 12.06 & 92\% & 118\%\\
\hline
\end{tabular}
\end{center}
\end{table}

\subsubsection{Full Image.}
We train a classifier on the full images, considering all 15 classes as well as empty images (16 total classes). On the cis-location test set we achieve a top-1 error of $20.83\%$, and a top-1 error of $41.08\%$ on the trans-location test set with a $97\%$ cis-to-trans increase in error. To investigate if requiring the classifier to both detect and classify animals increased overfitting on the training location backgrounds, we removed the empty images and retrained the classifiers using just the 15 animal classes. Performance stayed at nearly the same levels, with a top-1 error of $19.06\%$ and $41.04\%$ for cis- and trans-locations respectively. Utilizing sequence information helped reduce overall error (achieving errors of $14.92\%$ and $28.69\%$ on cis- and trans-locations respectively), but even in the most optimistic oracle setting, there is still a $92\%$ increase in error between evaluating on cis- and trans-locations. See Table~\ref{table:results} for the full results.

\subsubsection{Bounding Boxes.}
We train a classifier on cropped bounding boxes, excluding all empty images (as there is no bounding box in those cases). Using no sequence information we achieve a cis-location top-1 error of $8.14\%$ and a trans-location top-1 error of $19.56\%$. While the overall error has decreased compared to the image level classification, the error increase between cis- and trans-locations is still high at $140\%$. Sequence information further improved classification results (achieving errors of $5.52\%$ and $12.06\%$ on cis- and trans-locations respectively), and slightly reduced generalization error, bringing the error increase down to $118\%$ in the most optimistic setting. See Table~\ref{table:results} for the full results. Additional experiments investigating the effect of number of images per location, number of training locations, and selection of validation location can be seen in the supplementary material.

\subsubsection{Analysis}
\begin{figure}
\begin{minipage}[b]{.48\linewidth}
  \centering
  \centerline{\includegraphics[width=6.5cm]{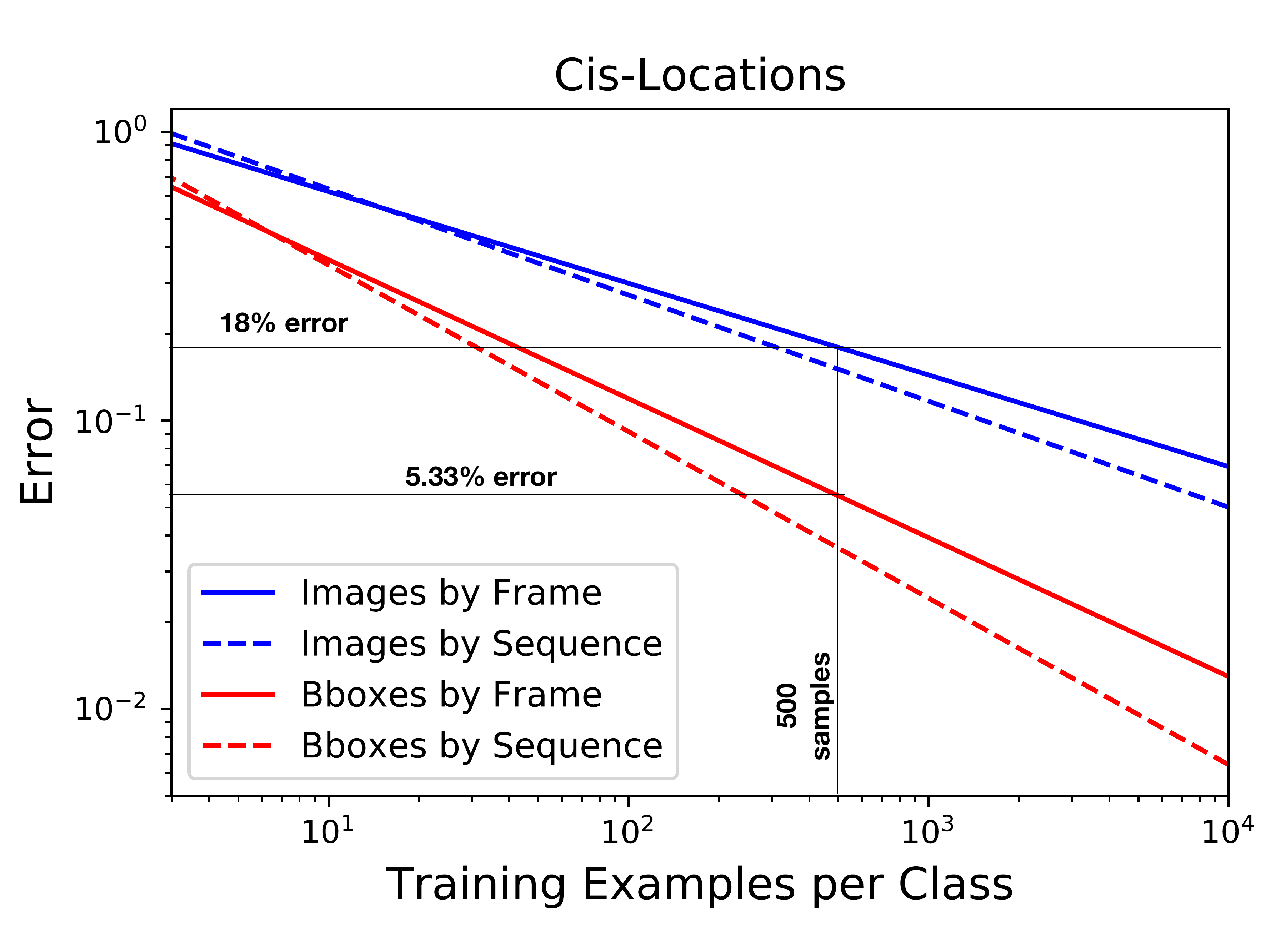}}
  
\end{minipage}
\hfill
\begin{minipage}[b]{0.48\linewidth}
  \centering
  \centerline{\includegraphics[width=6.5cm]{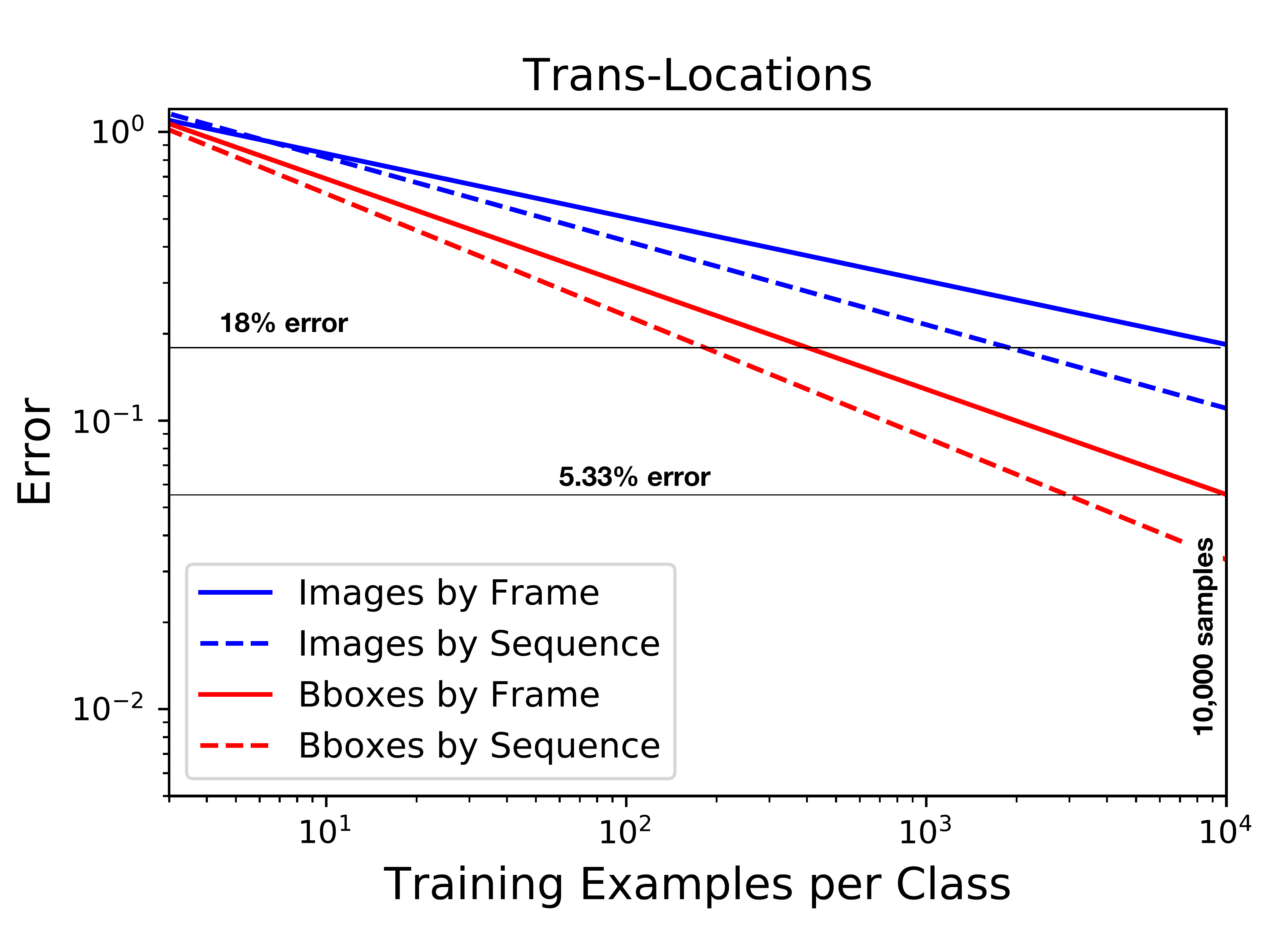}}
  
\end{minipage}
\caption{\textbf{Classification error vs. number of class-specific training examples}. Error is calculated as 1 - AUC (area under the precision-recall curve). Best-fit lines through the error-vs-n.examples points for each class in each scenario (points omitted for clarity), with average $r^2$ = 0.261. An example of line fit on top of data can be seen in Fig.~\ref{fig:specDist}. As expected, error decreases as a function of the number of training examples. This is true both for image classification (blue) and bounding-box classification (red) on both cis-locations and trans-locations. However, trans-locations show significantly higher error rates. To operate at an error rate of 5.33\% on bounding boxes or 18\% on images at the cis-locations we need 500 training examples, while we need  10,000 training examples to achieve the same error rate at the trans-locations, a 20x increase in data.}
\label{fig:samevnew}
\begin{minipage}[b]{.3\linewidth}
  \centering
  \centerline{\includegraphics[width=4.0cm]{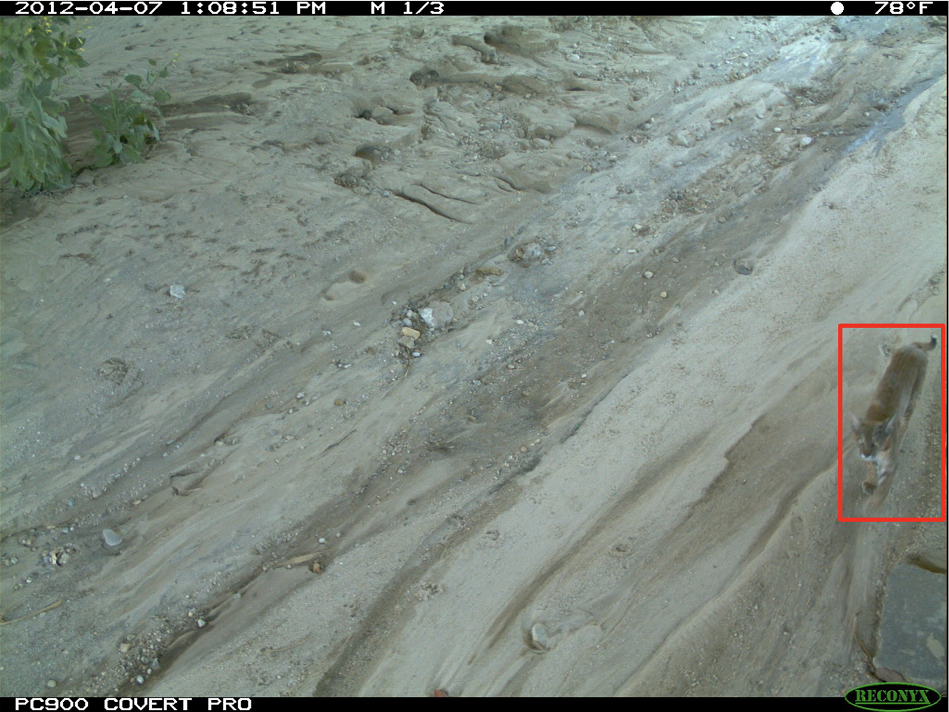}}
  \vspace{0.1in}
\end{minipage}
\hfill
\begin{minipage}[b]{0.3\linewidth}
  \centering
  \centerline{\includegraphics[width=4.0cm]{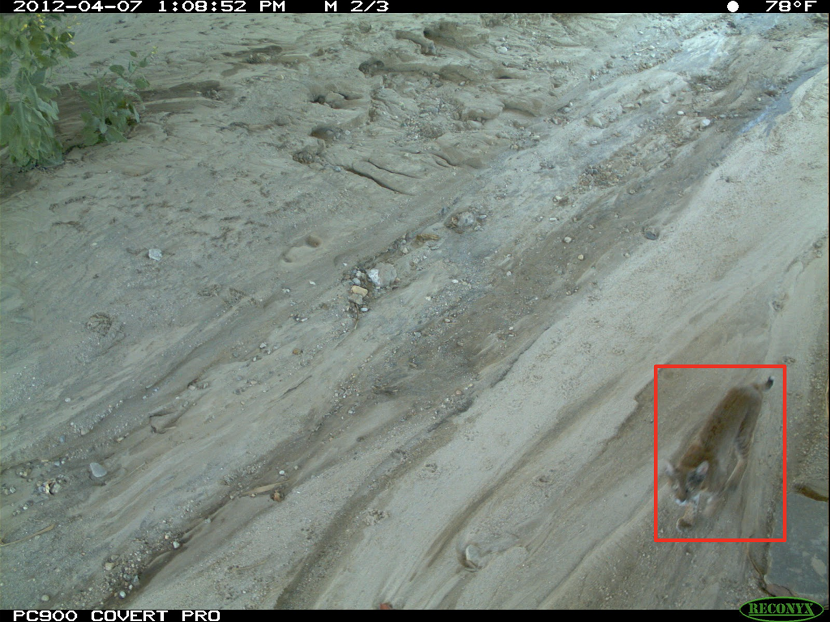}}
  \vspace{0.1in}
\end{minipage}
\hfill
\begin{minipage}[b]{.3\linewidth}
  \centering
  \centerline{\includegraphics[width=4.0cm]{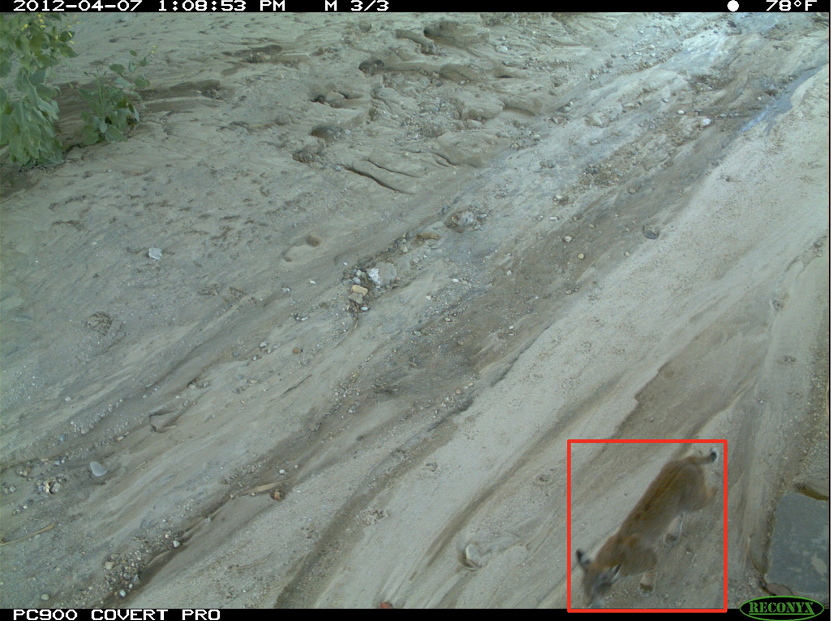}}
  \vspace{0.1in}
\end{minipage}
\begin{minipage}[b]{0.3\linewidth}
  \centering
  \centerline{\includegraphics[width=4.0cm]{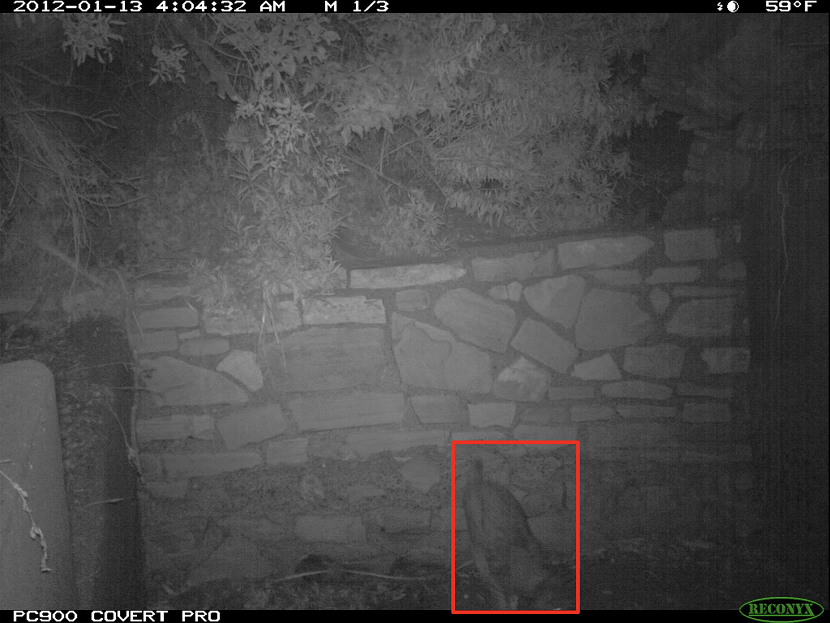}}
\end{minipage}
\hfill
\begin{minipage}[b]{.3\linewidth}
  \centering
  \centerline{\includegraphics[width=4.0cm]{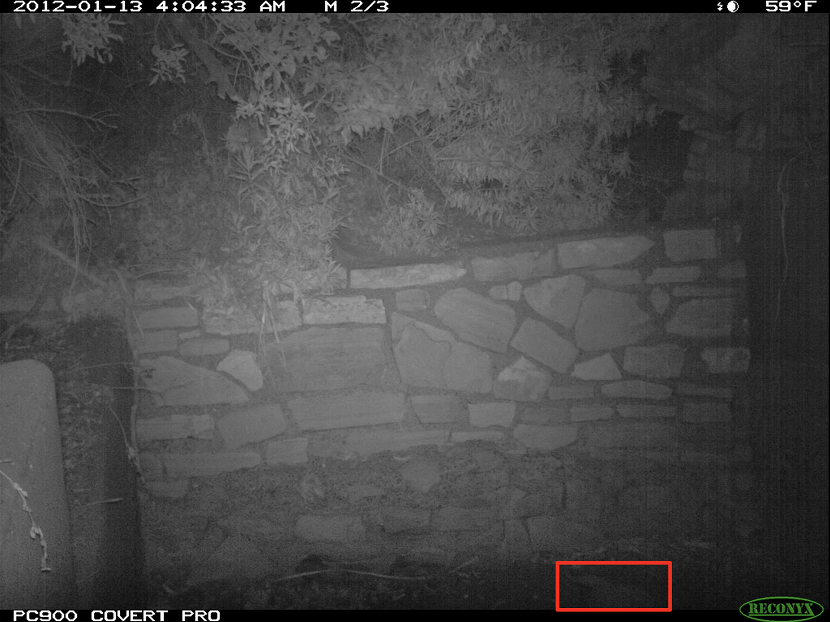}}
\end{minipage}
\hfill
\begin{minipage}[b]{0.3\linewidth}
  \centering
  \centerline{\includegraphics[width=4.0cm]{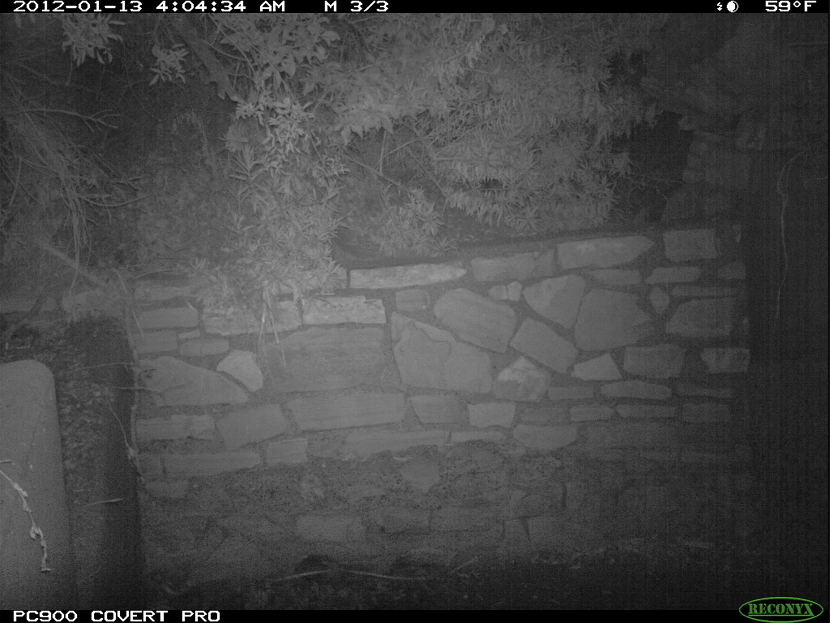}}
\end{minipage}

\caption{\textbf{Trans-classification failure cases at the sequence level}: (Based on classification of bounding box crops) In the first sequence, the network struggles to distinguish between `cat' and `bobcat', incorrectly predicting `cat' in all three images with a mean confidence of 0.82. In the second sequence, the network struggles to classify a bobcat at an unfamiliar pose in the first image and instead predicts `raccoon' with a confidence of 0.84. Little additional sequence information is available in this case, as the next frame contains only a blurry tail, and the last frame is empty}
\label{fig:failureCases}
\end{figure}

\begin{figure}
\centering
\begin{minipage}[b]{.48\linewidth}
  \centering
  \centerline{\includegraphics[width=6cm]{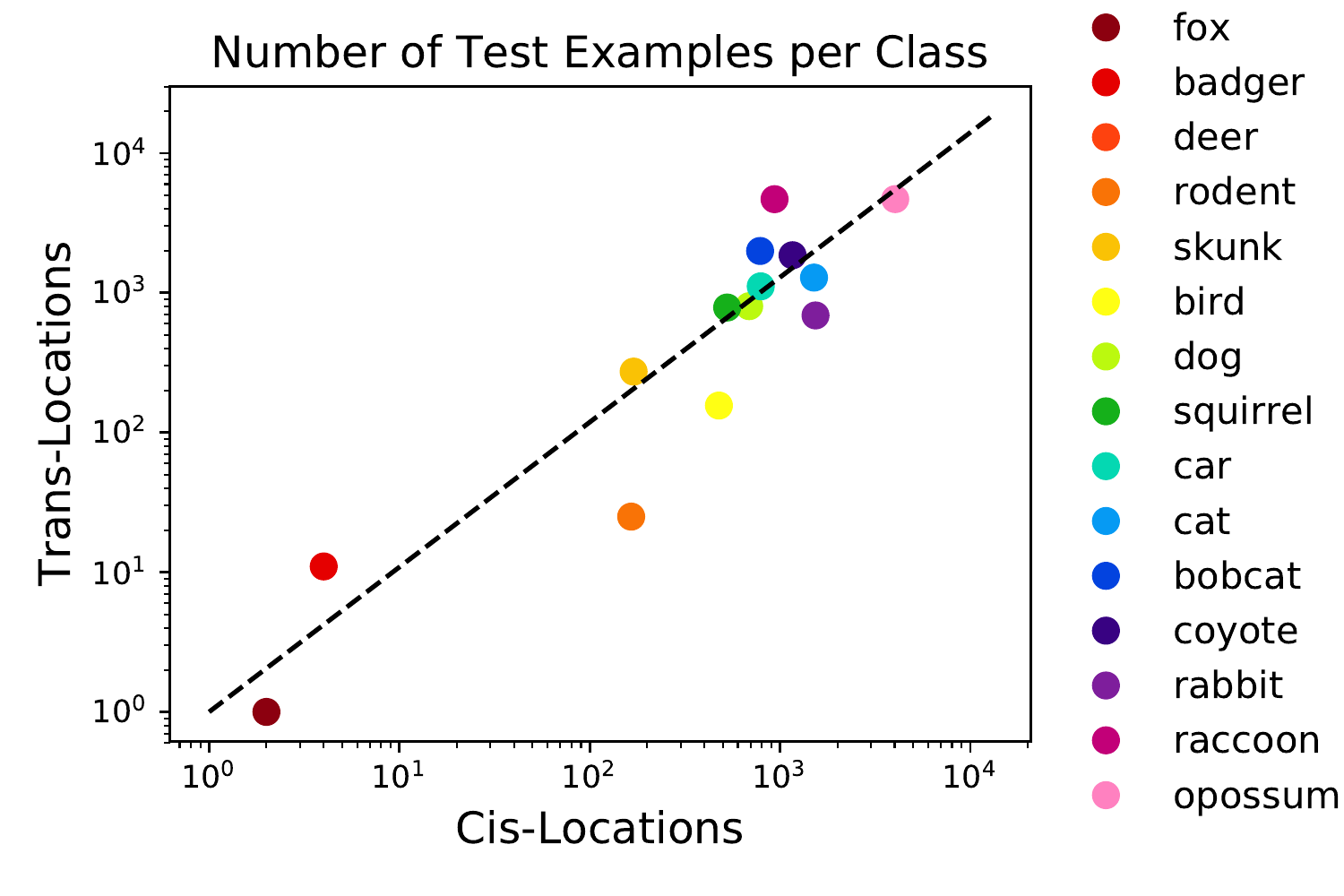}}
  
\end{minipage}
\hfill
\begin{minipage}[b]{0.48\linewidth}
  \centering
  \centerline{\includegraphics[width=6cm]{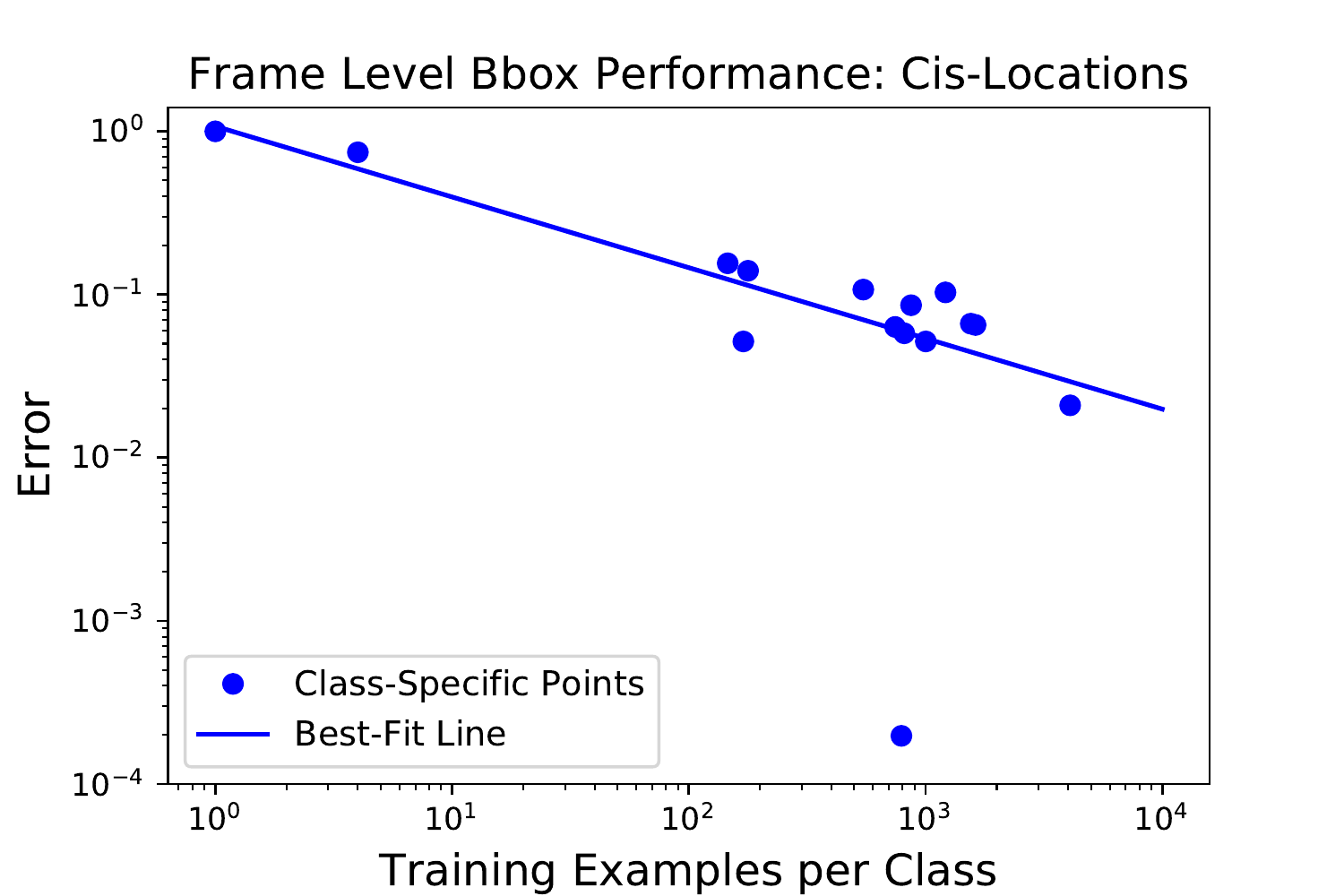}}
  \vspace{2mm}
  
\end{minipage}
\caption{(Left) Distribution of species across the two test sets. (Right) An example of line fit used to generate the plots in Fig.~\ref{fig:samevnew} }
\label{fig:specDist}
\end{figure}

Fig.~\ref{fig:samevnew} provides a high level summary of our experimental findings. Namely, there is a generalization gap between cis- and trans-locations. Cropped boxes help to improve overall performance (shifting the blue lines vertically downward to the red lines), but the gap remains. In the best case scenario (red dashed lines: cropped boxes and optimistically utilizing sequences) we see a $92\%$ increase in error between the cis- and trans-locations (with the same number of training examples), and 20x increase in training examples to have the same error rate. One might wonder whether this generalization gap is due to a large shift in class distributions between the two locations types. However, Fig.~\ref{fig:specDist} shows that the overall distribution of classes between the locations is similar, and therefore probably does not account for the performance loss. 

\subsection{Detection}

We use the Faster-RCNN implementation found in the Tensorflow Object Detection code base \cite{huang2017speed} as our detection model. We study performance of the Faster-RCNN model using two different backbones, ResNet-101 \cite{he2016deep} and Inception-ResNet-v2 with atrous convolution \cite{huang2017speed}. Similar to our classification experiments we analyze the effects of using sequence information using two methods:  \textbf{(1) Most Confident} we consider a sequence to be labeled correctly if the most confident detection across \textit{all} frames has an IoU$ \geq 0.5$ with its matched ground truth box; \textbf{(2) Oracle} we consider a sequence to be labeled correctly if \textit{any} frame's most confident detection has IoU$ \geq 0.5$ with its matched ground truth box. Note that method (2) is more optimistic than method (1).

Our detection models are pretrained on COCO \cite{lin2014microsoft}, images are resized to have a max dimension of 1024 and a minimum dimension of 600; each experiment uses SGD with a momentum of 0.9 and a fixed learning rate schedule. Starting at $0.0003$ we decay the learning rate by a factor of $10$ at $90$k steps and $120$k steps. We use a batch size of 1, and employ horizontal flipping for data augmentation. For evaluation, we consider a detected box to be correct if its IoU$ \geq 0.5$ with a ground truth box.   

Results from our experiments can be seen in Table~\ref{table:detect_results} and Fig~\ref{fig:detectPR}. We find that both backbone architectures perform similarly. Without taking sequence information into account, the models achieve $\sim77\%$ mAP on cis-locations and $\sim71\%$ mAP on trans-locations. Adding sequence information using the most confident metric improves results, bringing performance on cis- and trans-locations to similar values at $\sim85\%$. Finally, using the oracle metric brings mAP into the $90$s for both locations. Precision-recall curves at the frame and sequence levels for both detectors can be seen in Fig.~\ref{fig:detectPR}.

\subsubsection{Analysis}
There is a significantly lower generalization error in our detection experiments when not using sequences than what we observed in the classification experiments ($\sim30\%$ error increase for detections vs $\sim115\%$ error increase for classification). When using sequence information, the generalization error for detections is reduced to only $\sim5\%$.

\begin{figure}
\begin{minipage}[b]{.3\linewidth}
  \centering
  \centerline{\includegraphics[width=4.0cm]{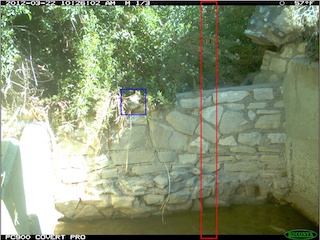}}
  \vspace{0.1in}
\end{minipage}
\hfill
\begin{minipage}[b]{0.3\linewidth}
  \centering
  \centerline{\includegraphics[width=4.0cm]{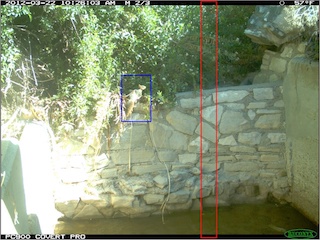}}
\vspace{0.1in}
\end{minipage}
\hfill
\begin{minipage}[b]{.3\linewidth}
  \centering
  \centerline{\includegraphics[width=4.0cm]{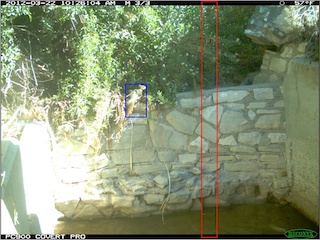}}
  \vspace{0.1in}
\end{minipage}
\begin{minipage}[b]{0.3\linewidth}
  \centering
  \centerline{\includegraphics[width=4.0cm]{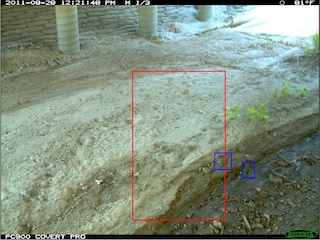}}
  \vspace{0.1in}
\end{minipage}
\hfill
\begin{minipage}[b]{.3\linewidth}
  \centering
  \centerline{\includegraphics[width=4.0cm]{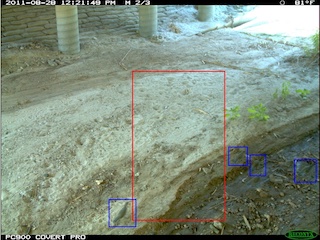}}
  \vspace{0.1in}
\end{minipage}
\hfill
\begin{minipage}[b]{0.3\linewidth}
  \centering
  \centerline{\includegraphics[width=4.0cm]{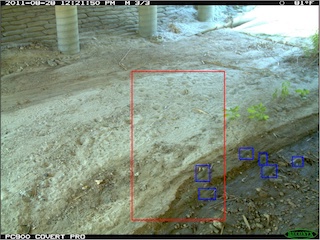}}
  \vspace{0.1in}
\end{minipage}
\begin{minipage}[b]{0.3\linewidth}
  \centering
  \centerline{\includegraphics[width=4.0cm]{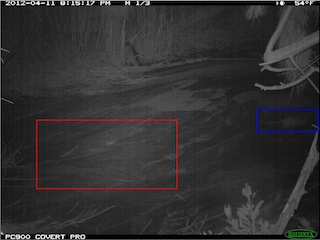}}
\end{minipage}
\hfill
\begin{minipage}[b]{.3\linewidth}
  \centering
  \centerline{\includegraphics[width=4.0cm]{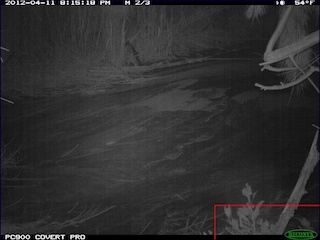}}
\end{minipage}
\hfill
\begin{minipage}[b]{0.3\linewidth}
  \centering
  \centerline{\includegraphics[width=4.0cm]{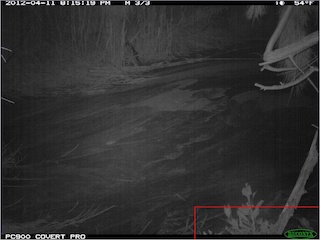}}
\end{minipage}

\caption{\textbf{Trans-detection failure cases at the sequence level}: Highest-confidence detection in red, ground truth in blue. In all cases the confidence of the detection was lower than 0.2. The first two sequences have small ROI, compounded with challenging lighting in the first and camouflaged birds in the second. In the third the opossum is poorly illuminated and only visible in the first frame.}
\label{fig:detectFailureCases}
\centering
\begin{minipage}[b]{.48\linewidth}
  \centering
  \centerline{\includegraphics[width=7cm]{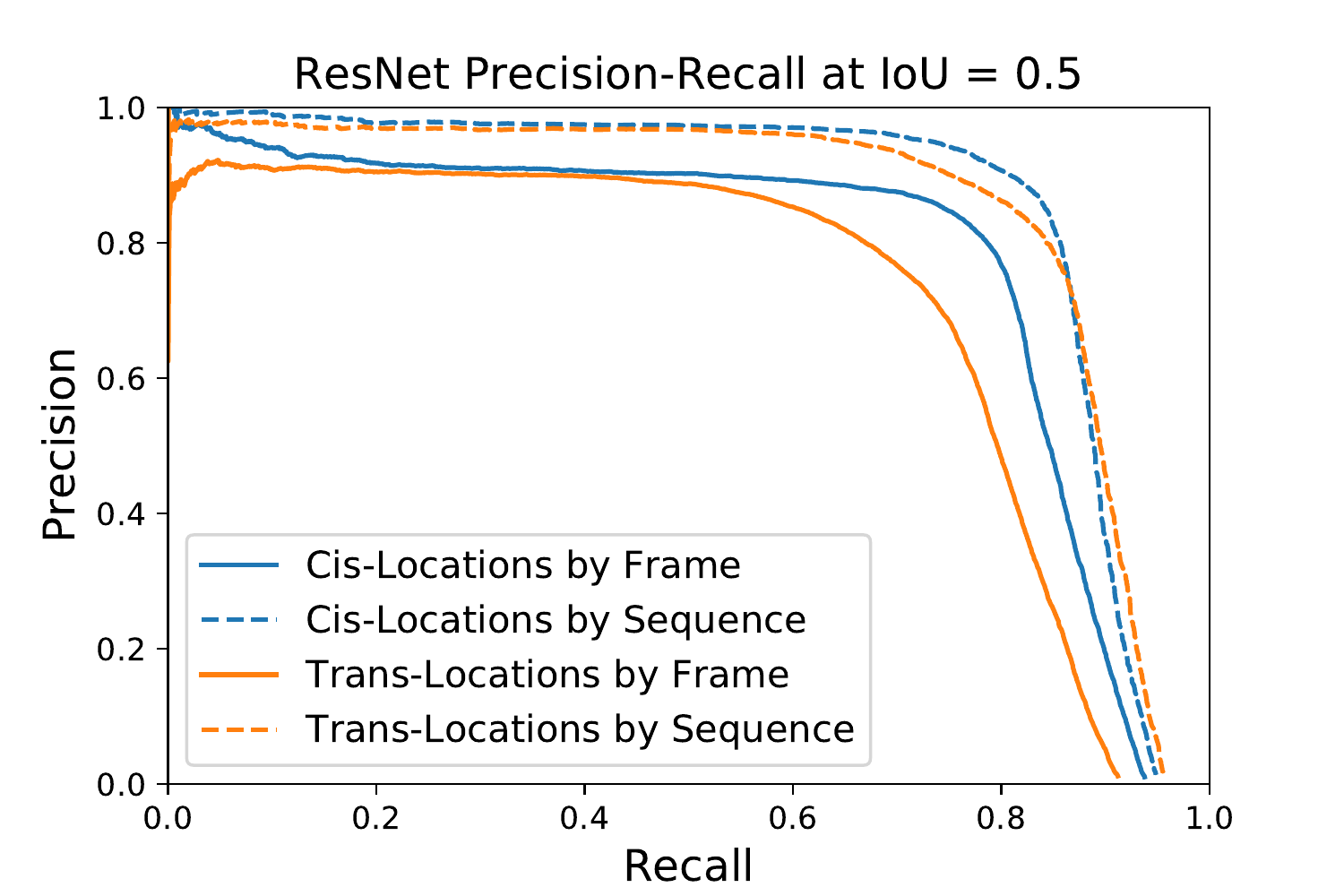}}
  
\end{minipage}
\hfill
\begin{minipage}[b]{0.48\linewidth}
  \centering
  \centerline{\includegraphics[width=7cm]{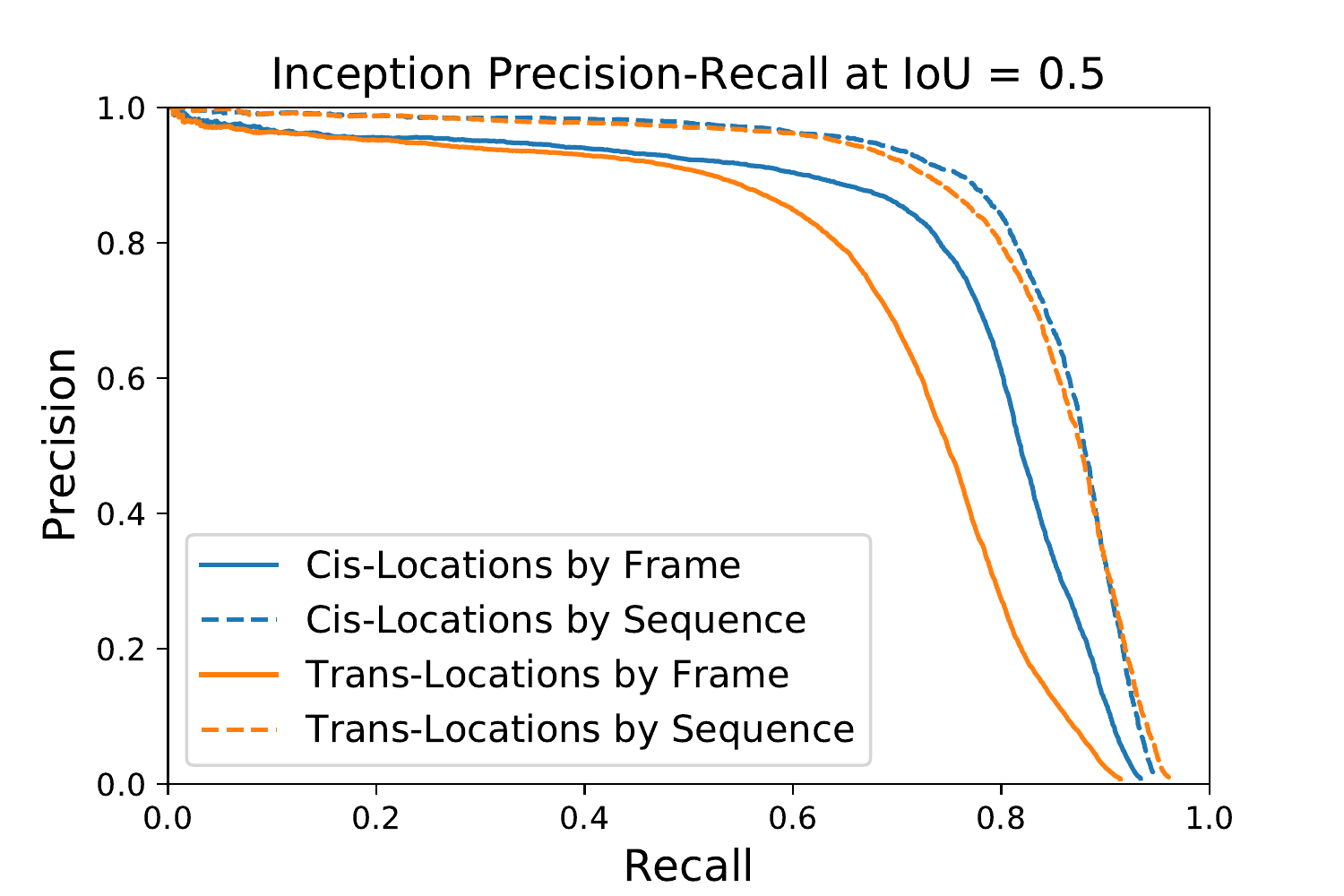}}
  
\end{minipage}
\caption{Faster-RCNN precision-recall curves at an IoU of 0.5, by frame and by sequence, using a confidence-based approach to determine which frame should represent the sequence}
\label{fig:detectPR}
\end{figure}

Qualitatively, we found the mistakes can often be attributed to nuisance factors that make frames difficult. We see examples of all 6 nuisance factors described in Fig.~\ref{fig:challenging_ims} causing detection failures. The errors remaining at the sequence level occur when these nuisance factors are present in all frames of a sequence, or when the sequence only contains a single, challenging frame containing an animal. Examples of sequence-level detection failures can be seen in Fig.~\ref{fig:detectFailureCases}. The generalization gap at the frame level implies that our models are better able to deal with nuisance factors at locations seen during training.   

Our experiments show that there is a small generalization gap when we use sequence information. However, overall performance has not saturated, and current state-of-the-art detectors are not achieving high precision at high recall values ($1\%$ precision at recall$=95\%$). So while we are encouraged by the results, there is still room for improvement. When we consider frames independently, we see that the generalization gap reappears. Admittedly this is a difficult case as it is not clear what the performance of a human would be without sequence information. However, we know that there are objects that can be detected in these frames and this dataset will challenge the next generation of detection models to accurately localize these difficult cases.

\setlength{\tabcolsep}{4pt}
\begin{table}
\begin{center}
\caption{Detection mAP at IoU=0.5 across experiments.}
\label{table:detect_results}
\begin{tabular}{l|cc|cc|cc|}
\cline{2-7}
&  \multicolumn{2}{|c|}{Cis-Locations} & \multicolumn{2}{|c|}{Trans-Locations} & \multicolumn{2}{|c|}{Error Increase}\\
\hline
\multicolumn{1}{|c|}{Sequence Information}& ResNet & Inception & ResNet & Inception & ResNet & Inception\\
\hline
\multicolumn{1}{|c|}{None} & 77.10 & 77.57 & 70.17 & 71.37 & 30\% & 27.6\% \\
\multicolumn{1}{|c|}{Most Confident} & 84.78 & 86.22 & 84.09 & 85.44 & 4.5\%& 5.6\%\\
\multicolumn{1}{|c|}{Oracle}  & 94.95 & 95.04 & 92.13 & 93.09 & 55.8\% & 39.3\%\\
\hline
\end{tabular}
\end{center}
\end{table}

\section{Conclusions}

The question of generalization to novel image statistics is taking center stage in visual recognition. Many indicators point to the fact that current systems are data-inefficient and do not generalize well to new scenarios. Current systems are, in essence, glorified pattern-matching machines, rather than intelligent visual learners.

Many problem domains face a generalization challenge where the test conditions are potentially highly different than what has been seen during training. Self driving cars navigating new cities, rovers exploring new planets, security cameras installed in new buildings, and assistive technologies installed in new homes are all examples where good generalization is critical for a system to be useful. However, the most popular detection and classification benchmark datasets \cite{imagenet_cvpr09,lin2014microsoft,everingham2010pascal,openimages} are evaluating models on test distributions that are the same as the train distributions. Clearly it is important for models to do well on data coming from the same distribution as the train set. However, we argue that it is important to characterize the generalization behavior of these models when the test distribution deviates from the train distribution. Current datasets do not allow researchers to quantify the generalization behavior of their models. 

We contribute a new dataset and evaluation protocol designed specifically to analyze the generalization behavior of classification and detection models. Our experiments reveal that there is room for significant improvement on the generalization of state-of-the-art classification models. Detection helps to improve overall classification accuracy, and we find that while detectors generalize better to new locations, there is room to improve their precision at high recall rates.   

Camera traps provide a unique experimental setup that allow us to explore the generalization of models while controlling for many nuisance factors. Our current dataset is already revealing interesting behaviors of classification and detection models. There is still more information that we can learn by expanding our dataset in both data quantity and evaluation metrics. We plan to extend this dataset by adding additional locations, both from the American Southwest and from new regions. Drastic landscape and vegetation changes will allow us to investigate generalization in an even more challenging setting. Rare and novel events are frequently the most important and most challenging to detect and classify, and while our dataset already has these properties, we plan to define experimental protocols and data splits for benchmarking low-shot performance and the open-set problem of detecting and/or classifying species not seen during training.
\section{Acknowledgements}
We would like to thank the USGS and NPS for providing data. This work was supported by NSFGRFP Grant No. 1745301, the views are those of the authors and do not necessarily reflect the views of the NSF. Compute time was provided by an AWS Research Grant.

\bibliographystyle{splncs}
\bibliography{bibliography}
\newpage
\begin{center}
\large{\textbf{Recognition in Terra Incognita: Supplementary Material}}
\end{center}

\Section{Additional Experiments}
\begin{figure}
\begin{minipage}{0.48\linewidth}
  \centering
  \centerline{\includegraphics[width=6cm]{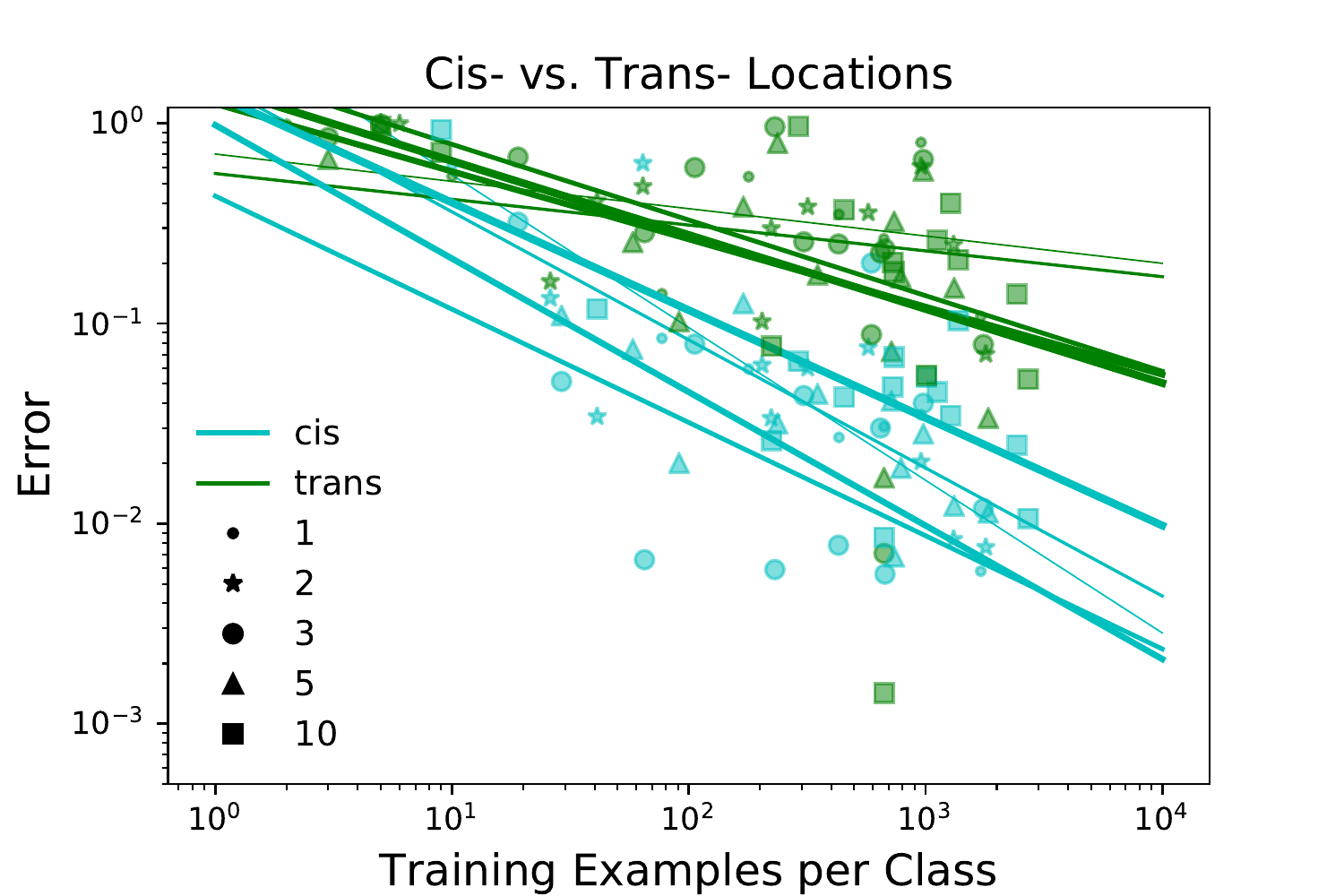}} 
\end{minipage}
\begin{minipage}{0.48\linewidth}
  \centering
  \centerline{\includegraphics[width=6cm]{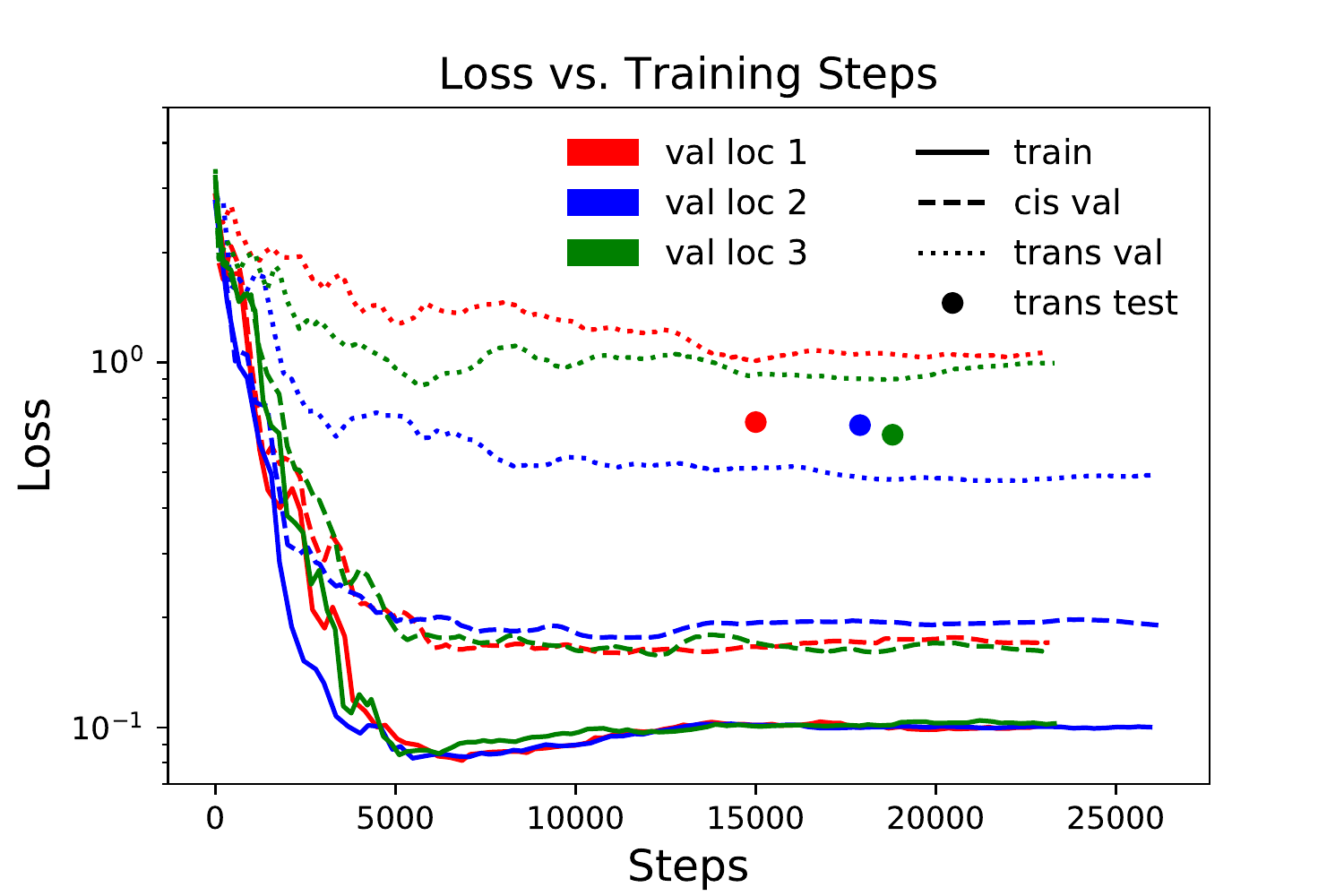}} 
\end{minipage}
\Caption{\textbf{Generalization metrics are robust to N. locations and to validation.} Both plots are based on bounding box classification. \textbf{(Left)} Error per class vs. number of training examples (best-fit line width denotes number of training locations in $1,2,3,5,10$). Trans performance is stable for $N$ locations with $2 < N \leq 10$. We chose 10 training locations to study generalization behaviors while providing maximal data for experimentation.
\textbf{(Right)} Loss curves using different locations for trans-validation. The test loss for the selected model for each validation set remains stable, implying that the choice of validation location does not greatly impact trans test performance.}
\label{fig:rebuttal}
\end{figure}

\Subsection{Varying the amount of training data per location}
We chose to use a small number of locations as this is a key variable of the generalization problem. In the limit, we would study the behavior of models trained on a single location with ``unlimited'' training data. We did not have access to such a dataset, and therefore used 10 training locations in order to have a sufficient number of training examples. To verify whether 10 training locations would yield significantly different results than 1 training location, we ran our bounding box experiments with a quarter, half, and all of the images available per training location, and saw trans test accuracies of $80.6\%$, $83.0\%$, and $83.4\%$ respectively. This implies that {\em increasing the number of images per location would not solve the generalization problem}.

\Subsection{Varying the number of training locations}
As an additional control, we experimented with varying the number of training locations (see Fig. \ref{fig:rebuttal}(Left)), and find that trans performance is stable as the number of training locations is increased beyond 2. Thus, we are confident that our dataset is adequate to measure generalization ability. We expect the generalization gap to narrow with $N >> 10$, but as the number of training locations increases the focus of the experiment shifts.  We want to provide a test bed to specifically study generalization when provided with few training locations.

\Subsection{Varying the validation location}
To analyze the effect of the validation split, we repeated our experiments with 2 other validation locations (see Fig. \ref{fig:rebuttal}(Right)). We find that test performance is relatively stable regardless of the validation split. Fig. \ref{fig:rebuttal}(Right) also shows training and validation curves for the three different validation experiments.

\begin{figure}
\centering
\includegraphics[height=10cm]{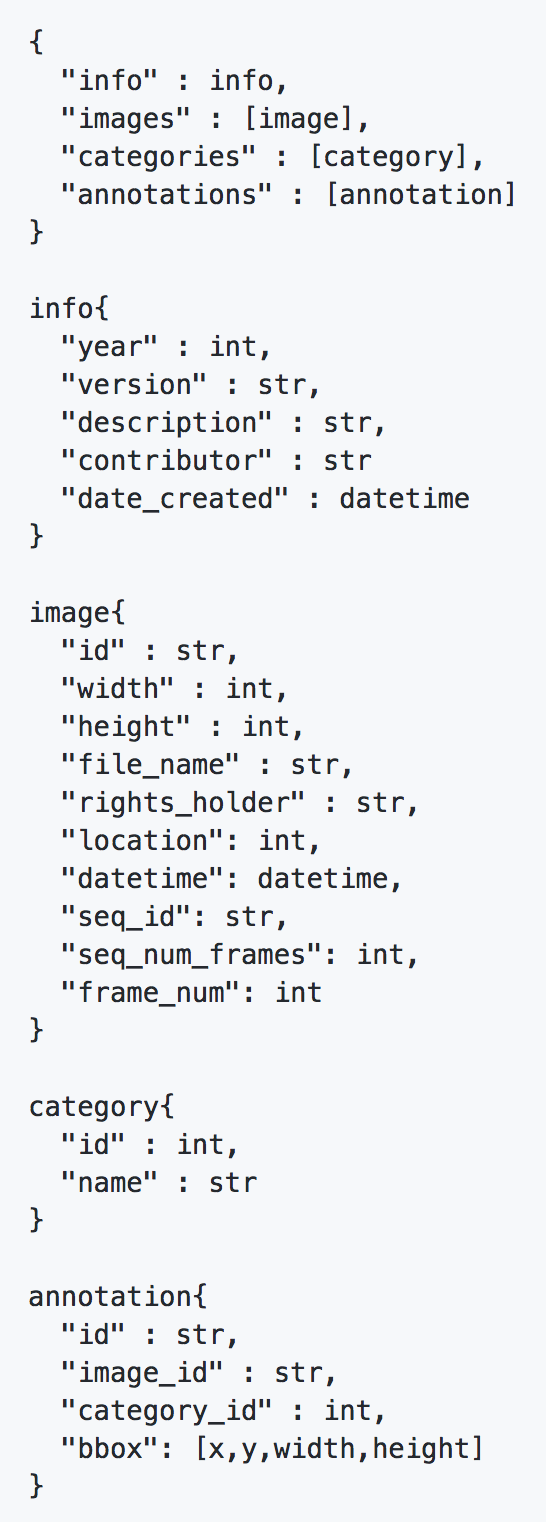}
\Caption{COCO-CameraTraps data format}
\label{fig:format}
\end{figure}

\Section{Data Format}
We chose to use an adapted version of the JSON format used by the COCO dataset with additional camera trap-specific fields, which we call COCO-CameraTraps. The format can be seen in Fig. \ref{fig:format}.

We added several fields for each image in order to specify camera-trap specific information.  These fields include a location id, a sequence id, the number of frames in that sequence, and the frame number of the individual image.  Note that not all cameras take sequences of images at a single trigger, so for some images the number of frames in the associated sequence will be one.

All data can be accessed at \url{https://beerys.github.io/CaltechCameraTraps/}.

\end{document}